\documentclass{article}
\usepackage[T1]{fontenc}
\usepackage{hyperref}
\usepackage[utf8]{inputenc}
\usepackage{aux/spconf}
\usepackage{amsmath,graphicx}
\usepackage{lipsum}
\usepackage{caption}
\usepackage{gensymb}
\usepackage{threeparttable}
\usepackage{tabularx}
\usepackage{booktabs}
\usepackage[noadjust]{cite}
 
\usepackage{filecontents}
\let\OLDthebibliography\thebibliography
\renewcommand\thebibliography[1]{
  \OLDthebibliography{#1}
  \setlength{\parskip}{0pt}
  \setlength{\itemsep}{0pt plus 0.3ex}
}

\title{Leveraging Adaptive Color Augmentation in Convolutional \\ Neural Networks for Deep Skin Lesion Segmentation}

\name{
Anindo Saha, 
Prem Prasad, 
Abdullah Thabit
}

\address{
Escola Politècnica Superior, Universitat de Girona, Spain \\
}

\begin{document}
\maketitle

\begin{abstract}
Fully automatic detection of skin lesions in dermatoscopic images can facilitate early diagnosis and repression of malignant melanoma and non-melanoma skin cancer. Although convolutional neural networks are a powerful solution, they are limited by the illumination spectrum of annotated dermatoscopic screening images, where color is an important discriminative feature. In this paper, we propose an adaptive color augmentation technique to amplify data expression and model performance, while regulating color difference and saturation to minimize the risks of using synthetic data. Through deep visualization, we qualitatively identify and verify the semantic structural features learned by the network for discriminating skin lesions against normal skin tissue. The overall system achieves a Dice Ratio of 0.891 with 0.943 sensitivity and 0.932 specificity on the ISIC 2018 Testing Set for segmentation.
\end{abstract}


\vspace{2mm}

\section{Introduction}
Human skin, as the largest organ of the body's integumentary system, is prone to a wide spectrum of cutaneous diseases and infections that can manifest as surface abnormalities or "lesions". In particular skin cancer, primarily non-melanoma skin cancer (NMSC) and the highly aggressive malignant melanoma (MM), represents the most common malignancy in Caucasians with over $1.3$ million new cases and $125,000$ deaths worldwide in 2018 \cite{1}. Early detection and diagnosis of skin lesions is critical to ensuring high survival rates \cite{2}. It can be achieved effectively, automatically and in real-time, by leveraging complex machine learning techniques and computer vision frameworks. With the emergence of large, multi-source dermatoscopic image datasets \cite{3} providing ample annotated training data, deep neural networks are now at the forefront of this technology.

In this research, we propose a deep convolutional neural network (CNN) to segment the most commonly occurring pigmented skin lesions. We incorporate a novel adaptive color augmentation technique, with improved functionality from its equivalent counterparts \cite{4,5}, to extend our training data representation. The augmentation exploits and accounts for the highly variable nature of dermatoscopic screening samples, where background illumination, hospital acquisition conditions and external obstructions can significantly modify the underlying color profile of a skin lesion captured in an image. Performance analysis is modelled after the \textit{ISIC 2018 Challenge: Skin Lesion Analysis for Melanoma Detection} \cite{6} using the HAM10000 \cite{3} dataset.

\section{Artificial Data Augmentation}
HAM10000 depicts $7$ types of skin lesions (melanoma, basal cell carcinoma, melanocytic nevus, actinic keratosis, benign keratosis, dermatofibroma, vascular lesion) in $10,015$ images. However, a complete visual representation demands a much larger, unfeasible number of images, i.e. theoretically every possible instance in nature. The most practical means of compensation is to anticipate and adapt near-realistic variations of the images beyond the pre-existing, limited dataset using data augmentation. This step is proven to have a significant impact on inference by reducing overfitting and improving generalization \cite{2}.

\subsection{Color Augmentation}
Color is an important feature for diagnosing melanoma since certain color markers are associated with different stages and classes of the disease \cite{2}. A popular approach to achieve color constancy is the \textit{Gray World} algorithm that assumes the average color in a given scene is achromatic, i.e. gray, and any deviations is caused due to effects of light sources \cite{7}. Based on this, the illuminant profile of an image can be estimated as the independent average intensities of its RGB color channels. In turn, the scaling factor for each channel ($\beta_R, \beta_G, \beta_B$) is its average intensity divided by that of all $3$ channels combined. Together, these scaling factors constitute as the illuminant scales ($\beta'$) required to transform any image to a certain illuminant profile ($\beta$), given as follows:

\begin{figure}[h!]
\centerline{\includegraphics[width=0.485\textwidth]{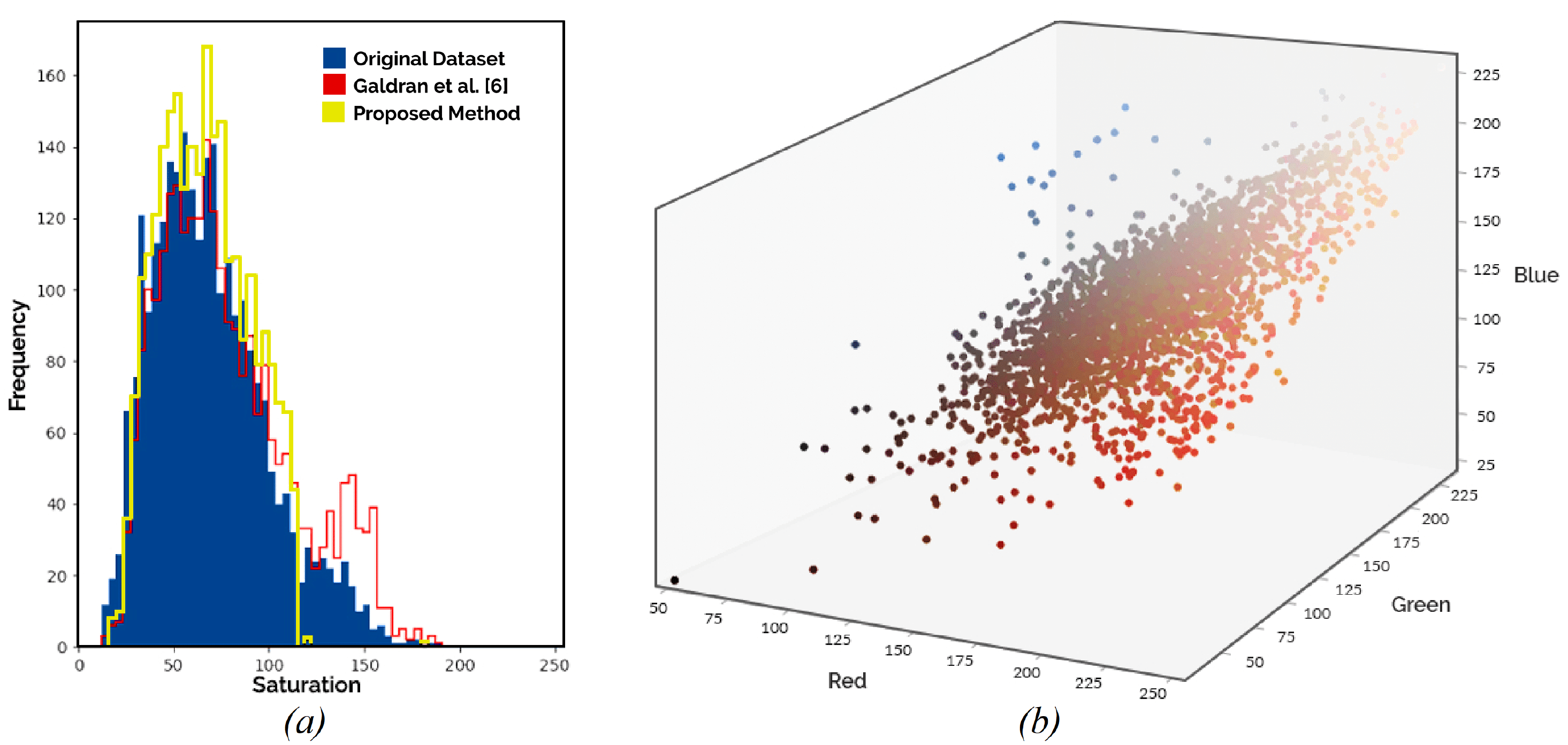}}
\captionsetup{belowskip=-8pt}
\caption{a) Histogram of mean saturation values across original training set and augmented datasets, derived from the coordinates of each image in the HSV color space. b) Distribution of illuminant profiles extracted from the training set by applying the \textit{Gray World} color constancy algorithm, where each marker displays its encoded color in the RGB color space.}
\label{fig1}
\end{figure}

\vspace{2mm}

\centerline{$\beta'\{\beta_R, \beta_G, \beta_B\} \equiv \beta'\{\frac{\frac{1}{n} \sum_{}{}I_{R}}{\frac{1}{3n} \sum_{}{} I_{RGB}}, \frac{\frac{1}{n} \sum_{}{} I_{G}}{\frac{1}{3n} \sum_{}{} I_{RGB}}, \frac{\frac{1}{n} \sum_{}{} I_{B}}{\frac{1}{3n} \sum_{}{} I_{RGB}}\}$}

\vspace{2mm}

Dividing the original image ($I$) by its illuminant scales ($\beta'$), results in a white-balanced image. Similarly, if we take the illuminant scales ($\gamma'$) of a different image and multiply it by the original white-balanced image, we obtain a color augmented image ($I_{aug}$) with the illuminant $\gamma$. 

\vspace{2mm}

\centerline{$I_{aug} = (I/\beta') * \gamma'  \hspace{3mm}$ }

\vspace{2mm}

This method was originally proposed by Galdran \textit{et al.} \cite{4}, but faces a number of important limitations. The authors perform color augmentations at train-time, casting each sample by an illuminant profile (with uniform probability distribution) that is randomly selected from the raw distribution of all illuminants present in the dataset. This leads to high variance while training the network, as a training set of $2072$ images can yield $2072^2$ possible variants, where each image is learned to different degrees, at random. In certain cases, augmented images can also exhibit an oversaturated, artificial hue \cite{4}, by adopting an illuminant profile vastly different from the original. These images can prove detrimental to the learning process and overwhelm the network. Hence, we propose an updated, adaptive approach to account for these limitations.

Fig. 1(b) illustrates how the overall illuminant distribution of the dataset is centered around a reddish hue, with a few outliers at both extremes. Saturation values follow accordingly, with a major Gaussian distribution centered around $65$ and a minor branch near $140$, as seen in Fig. 1(a). The secondary distribution represents highly saturated images in the original dataset and its proportion is boosted dramatically in a color augmented dataset generated by the original algorithm \cite{4}. Taking these factors into account, we introduce two strict conditions prior to augmentation: 

\vspace{2mm}

\centerline{$d(\beta, \gamma) = C * d(\beta, \alpha) \hspace{2mm} \wedge \hspace{2mm} S_\gamma \in [a,b]$ }

\vspace{2mm}

Here $d(\beta, \alpha)$ represents the Euclidean distance between the illumination profile ($\beta$) of an image ($I$) and its furthest counterpart ($\alpha$) in the RGB color space, i.e. their color difference. $d(\beta, \gamma)$ represents the equivalent between $\beta$ and the candidate illuminant to be used for augmentation ($\gamma$). $C$ is a thresholding factor used to control the color difference, empirically set as $0.4$. In other words, if $\gamma$ is sufficiently distinct $(40\%$ of maximum) from $\beta$, yielding new trainable data, but not as radically different as $\alpha$ (maximum possible color difference) such as to impair learning, we consider it as a positive candidate for augmentation. For the second condition, $S_\gamma$ represents the mean saturation value of $I_{aug}$ (post-augmentation with $\gamma$). If $S_\gamma$ pertains to the major distribution range $[a,b]$ of saturation in the original dataset ($a$ and $b$ are determined to be $15$ and $115$, respectively, for HAM10000), then we confirm the augmentation and use it for training. Otherwise, we iterate to the next candidate illuminant satisfying the first condition and verify the same. Fig. 2 illustrates this selection process. The process is repeated till both conditions are satisfied, upon which the algorithm effectively eliminates the generation of oversaturated, highly artificial images during augmentation. At train-time, the color augmented images are used alongside the original training set with a ratio of 1:1.

\begin{figure}[h!]
\centerline{\includegraphics[width=0.40\textwidth]{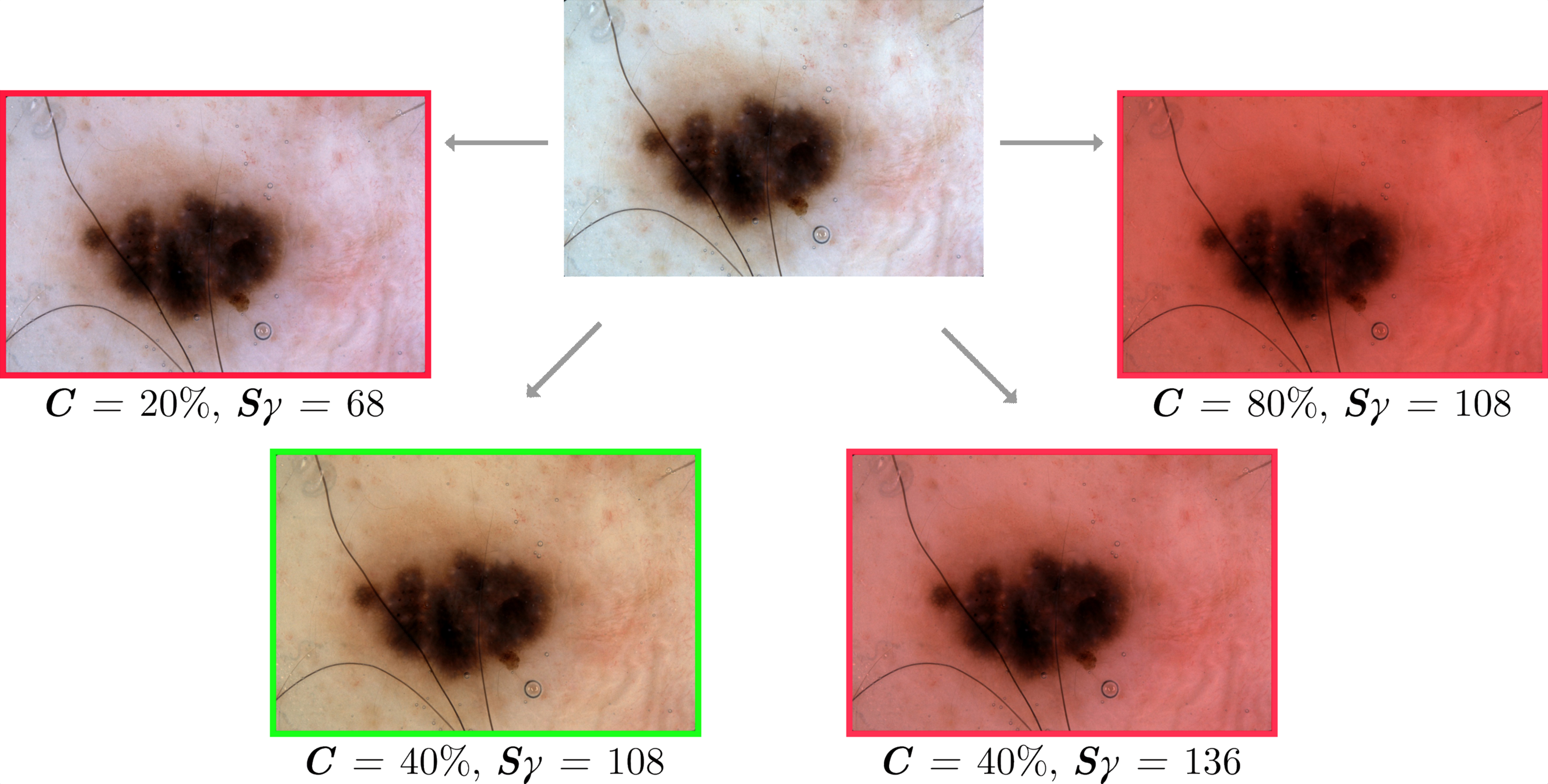}}
\captionsetup{belowskip=-10pt}
\caption{Possible color augmentations for an image and their respective values for $C$ and $S_\gamma$, with the negative candidates marked in red and the positive candidate (fulfilling both pre-requisite conditions) marked in green.}
\end{figure}

\subsection{Spatial Augmentation}
Morphological rigid transformations, such as rotation ($-180\degree$ to $180\degree$), flip and translation ($10\%$ along horizontal/vertical axis), are used to create spatial augmentations at train-time and account for learning orientations beyond the dataset.

\begin{figure*}[t!]
\centerline{\includegraphics[width=\textwidth]{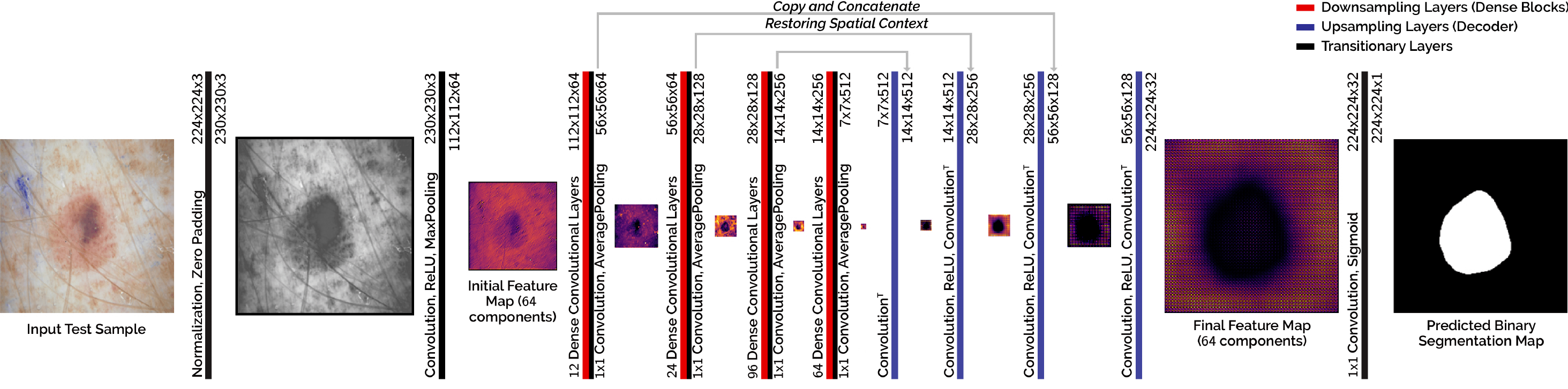}}
\caption{Architecture schematic for modified U-Net, with DenseNet201 backbone. Every dense block is made of several alternating $1\times1$ and $3\times3$ convolutional layer pairs, and considers diverse feature maps of all preceding layers.}
\label{fig2}
\end{figure*}

\vspace{4mm}

\section{Network Architecture}

The base architecture is a variant of the U-Net, as proposed by O. Ronneberger \textit{et al.} \cite{8} for biomedical image segmentation. It comprises of a backbone encoder (series of downsampling convolutional layers used for feature extraction) followed by a decoder (corresponding number of upsampling transposed convolutional layers) to deliver pixel-level classification in an output segmentation map of the original input size. All images are normalized and pre-processed to $224\times224$ pixels for ease of computation and uniformity through the tuning and optimization phase, and these dimensions serve as the input size for the network. Binary cross-entropy loss and Jaccard Index are used as the training metrics with gradient descent via Adam optimizer in backpropagation. To determine the most appropriate backbone for this network, a comparative analysis is drawn across several notable architectures (ResNet50, VGGNet16, VGGNet19, Inception-v3), each independently combined with the U-Net in separate turns. Due to its superior performance, DenseNet201 is selected as the encoder in the final model. DenseNet layers take the concatenated feature maps of all preceding layers as their input and similarly pass on their output feature map to all subsequent layers \cite{9}. As a result, the model requires less channels, is computationally/memory-wise efficient, has strong gradient flow during backpropagation and considers diverse features of different complexities at every stage of computation. The complete U-Net (as seen in Fig. 3) is then tuned across its hyperparameters and trained to maximize Jaccard Index over $5$-fold Monte Carlo cross-validation. Within the scope of this research, the optimal hyperparameters have been deduced as an exponentially decaying learning rate starting at $10$\textsuperscript{-4} ($50\%$ decay rate, patience of $2$ epochs) and a mini-batch size $16$ over a training period of $13$ epochs. The encoder is initialized with pre-trained ImageNet \cite{10} weights and the decoder follows Xavier initialization. Rectified Linear Unit (ReLU) is used to activate all convolutional layers.

\section{Results}
The  model  is  implemented using Keras API with TensorFlow as the backend engine, with a single NVIDIA Tesla K80 GPU used for hardware acceleration via Google Cloud Platform. After a final training run on the complete annotated \textit{ISIC 2018 Task 1: Training Set}, the segmentation results from inference on \textit{ISIC 2018 Task 1: Testing Set} have been recorded in Table 1. The same pipeline is followed for the corresponding datasets from the \textit{ISIC 2017 Challenge}. Post-processing includes simply smoothing and extracting the largest connected component from the predicted binary mask. Thresholded Jaccard Index, defined as the Jaccard Index of an image if it scores above $65\%$ and zero otherwise, is the primary evaluation metric for the 2018 edition of the challenge.

\vspace{3mm}

\begin{table}[!h]
\centering
\setlength{\tabcolsep}{2.05pt}
\renewcommand{\arraystretch}{1.0}
\captionsetup{belowskip=0pt}
\captionsetup{justification=centering}
\caption{Segmentation Performance for ISIC Testing Set (TJA: Thresholded Jaccard Index; JA: Jaccard Index; DI: Dice Ratio; AC: Accuracy; SE: Sensitivity; SP: Specificity)}
\begin{threeparttable}
\scalebox{1}{
\begin{tabular}{lccccc}
\toprule
Method  &  TJA  &  JA  &  DI  &  SE  &  SP\\
\midrule
Galdran et al. \cite{4}\textsuperscript{1}      
& $-$ & $0.767$ & $0.846$ & $0.865$ & $0.980$\\

Yuan et al. \cite{11}\textsuperscript{1}    
& $-$ & $0.765$ & $0.849$ & $0.825$ & $0.975$\\

\textbf{Proposed Model}\textsuperscript{1}      
& $-$ & $\boldsymbol{0.772}$ & $\boldsymbol{0.855}$ & $\boldsymbol{0.824}$ & $\boldsymbol{0.981}$\\

\hline

Shahin et al. \cite{12}\textsuperscript{2}      
& $0.738$ & $0.837$ & $0.903$ & $0.902$ & $0.974$\\

Bissoto et al. \cite{13}\textsuperscript{2}     
& $0.728$ & $0.792$ & $0.873$ & $0.934$ & $0.936$\\

\textbf{Proposed Model}\textsuperscript{2}       
& $\boldsymbol{0.771}$ & $\boldsymbol{0.819}$ & $\boldsymbol{0.891}$ & $\boldsymbol{0.943}$ & $\boldsymbol{0.932}$\\
\bottomrule
\end{tabular}}

\begin{tablenotes}
   \item[1]Training:Test Ratio = 2000:600 (ISIC 2017)
   \item[2]Training:Test Ratio = 2594:1000 (ISIC 2018)
\end{tablenotes}
\end{threeparttable}
\end{table}

\vspace{-3pt}

\section{Discussion} 
\subsection{Learning Effect of Color Augmentation}
Color augmentation is able to generate competitive segmentation results even with a relatively simple base model (without visual attention guidance, ensembling or application-driven custom layers). Although the entirety of the network benefits from a larger dataset, the broader spectrum of illumination profiles can also boost the performance of the augmentation algorithm. Furthermore, while competing methods are susceptible to a relatively lower sensitivity, models driven by color augmentation boast the highest sensitivity scores (as seen in Table 1). It should be noted that an alternate adaptation of the algorithm, operating in the perceptually uniform CIELAB color space, had also been implemented. Although it had proven to be computationally expensive, it yields no significant gain in segmentation performance. Finally, the current approach uses a single set of color augmented images, but a vast number of unique sets can be generated from the dataset. In a future iteration, we propose an ensemble network, where each sub-model is trained on an unique set of augmented images alongside the original training set, thereby utilizing more variations without overwhelming the network.

\subsection{Reusing Color-Based Segmentation Features}
By initializing the model backbone (DenseNet201) with pre-trained ImageNet weights \cite{10}, the network benefits from a strong understanding of fundamental features built upon $10$ million images, as well as faster convergence times. However, this step can be further extended. Once the U-Net is successfully trained to perform segmentation, we obtain a network that is highly responsive to discriminative skin lesion features with a notable sensitivity towards color. This can be verified by passing a white noise RGB image to the model and performing deep visualization \cite{14}. Colors arise due to shifts in each of the RGB channels by different amounts. By initializing an equivalent encoder for lesion classification using these new weights, we can relay the semantic structural information required to segment skin lesions from normal skin, as a foundation for the more complex features required to be learned in order to discriminate between different types of skin lesions $-$analogous to curriculum learning \cite{15}. A similar approach has been demonstrated by Mehta et al.\cite{16}, where the authors successfully illustrated the advantages of utilizing segmentation features for the classification of breast biopsy, using a singular stream of shared features from the backbone encoder. This remains an important area of interest to be investigated for skin lesion analysis.

\begin{figure}[h!]
\centerline{\includegraphics[width=0.485\textwidth]{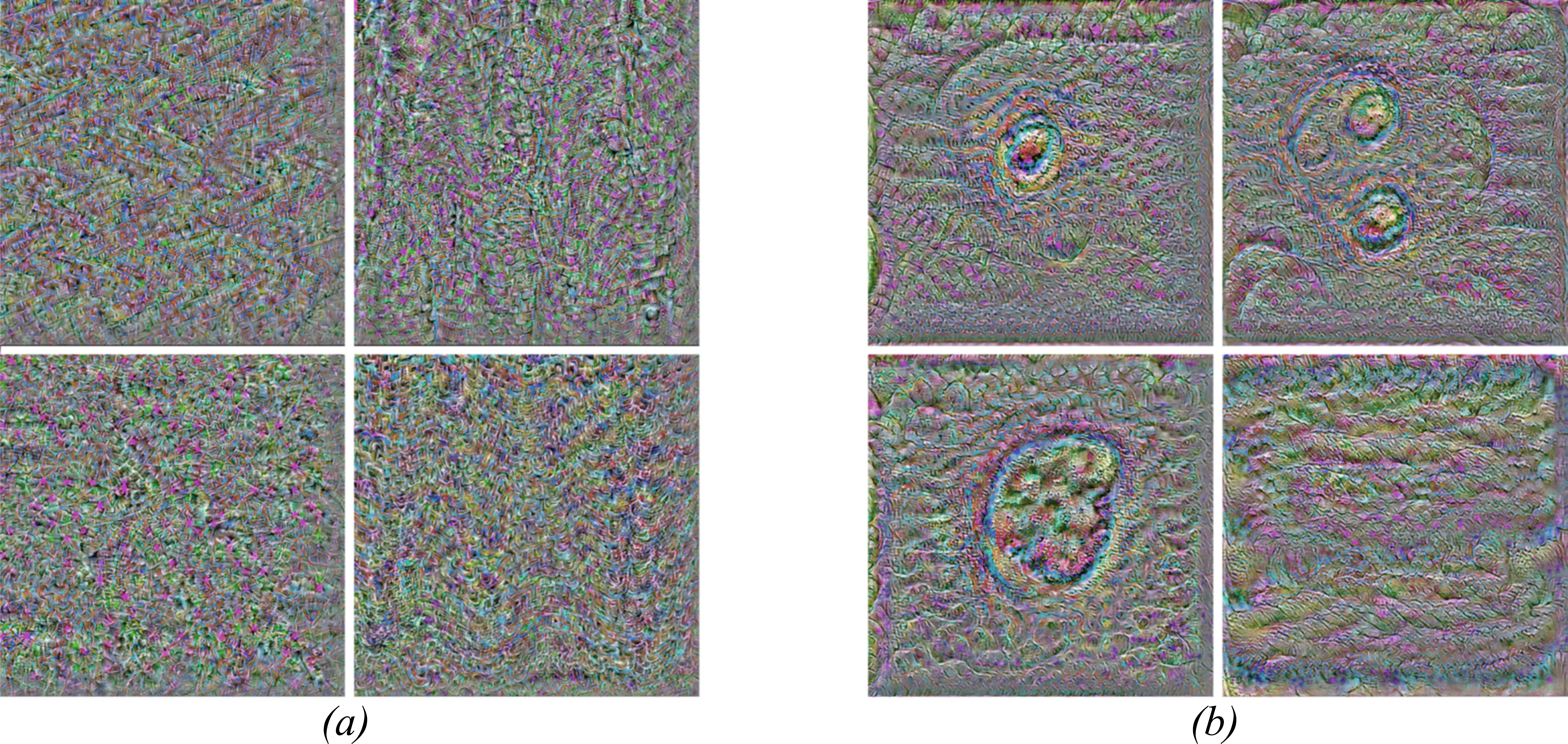}}
\captionsetup{belowskip=-10pt}
\caption{Deep visualization of filter activations in the final convolutional layer of the proposed U-Net, when trained on: a) ImageNet b) HAM10000. In the case of b), filter activations indicate more specific, higher-level patterns comparable to edge/texture features of skin lesions in dermatoscopic images.}
\end{figure}

\section{Conclusion}
In conclusion, adaptive color augmentation in deep convolutional neural networks has been redesigned for skin lesion analysis in dermatoscopic screening images. By considering color difference, saturation and the overall distribution of illumination profiles in the RGB color space, every augmentation is regulated to prevent oversaturation and the generation of artificial hues. The results are promising, verifying the importance of further research to draw out the full potential of color-based features in the role of automated, seamless diagnosis of melanoma and non-melanoma skin cancer.

\end{document}